\documentclass{article}

\PassOptionsToPackage{numbers, compress}{natbib}
\usepackage[preprint]{neurips_2026}

\usepackage[utf8]{inputenc}
\usepackage[T1]{fontenc}
\usepackage{hyperref}
\usepackage{url}
\usepackage{booktabs}
\usepackage{amsfonts}
\usepackage{amsmath}
\usepackage{nicefrac}
\usepackage{microtype}
\usepackage{xcolor}
\usepackage{graphicx}
\usepackage{subcaption}
\usepackage{multirow}
\usepackage{wrapfig}

\newcommand{\RR}{\mathbb{R}}

\title{Decompose Sparsely Where You Should, Absorb Densely Where You Should Not}

\author{%
  Ruixuan ~Deng\\
  School of Psychology \\
  Georgia Institute of Technology\\
  \texttt{rdeng62@gatech.edu} \\
  \And
  Zehao ~Jin\\
  College of Computing \\
  Georgia Institute of Technology\\
  \texttt{zehao@gatech.edu} \\
  \And
  Zekun ~Wang\\
  School of Interactive Computing \\
  Georgia Institute of Technology\\
  \texttt{zekun@gatech.edu} \\
  \And
  Zihan ~Dong\\
  College of Computing \\
  Georgia Institute of Technology\\
  \texttt{zdong312@gatech.edu} \\
}

\begin{document}
\maketitle

\begin{abstract}
Sparse autoencoders (SAEs) are typically trained to reconstruct the \textbf{entire} residual stream through a sparse dictionary, implicitly assuming that all activation content is amenable to sparse, monosemantic decomposition. We question this assumption and hypothesize that activations contain a low-rank, dense component that is computationally important to the model yet inherently unsuitable for sparse representation, which serves as a major source of the persistent dense latents widely observed in trained SAEs. To test this, we add a small rank-$r$ linear bottleneck in parallel with standard SAEs (BatchTopK and Matryoshka), allowing dense structure to be absorbed before sparse reconstruction. On Gemma-2-2B layer 12, a rank-24 bottleneck reduces dense latent count by up to 84\% while improving sparse probing and targeted probe perturbation on both architectures at matched sparsity. The absorbed component is (i) \textbf{structurally identifiable} as the top principal components and outlier dimensions; (ii) \textbf{causally necessary}, with removing it raising next-token cross-entropy by 7.5$\times$, far exceeding the 2.8$\times$ from removing the geometrically near-identical top-24 PCA directions; and (iii) \textbf{redundantly encoded by sparse dictionaries}, with ablating 787 maximally aligned sparse features raising cross-entropy by only 2.9$\times$ and ablating 2,048 topic-aligned features leaving MMLU topic classification virtually unchanged, whereas removing the scaffold drops it from 98.7\% to chance. Together, our findings identify a compact, semantically informative and causally important component of residual stream activations (which we term a \textbf{computational scaffold}) that standard sparse dictionaries represent inefficiently, suggesting that the scope of sparsity-based interpretability methods warrants careful re-examination.
\end{abstract}

%==============================================================================
\section{Introduction}
\label{sec:intro}
%==============================================================================
 
Sparse autoencoders (SAEs) have become the predominant tool for extracting interpretable features from language model activations~\citep{bricken2023monosemanticity, Cunningham2023a}, enabling mechanistic analysis of increasingly large models~\citep{templeton2024scaling}. The standard approach trains an SAE to reconstruct the \textbf{entire} residual stream activation $\mathbf{x} \in \RR^D$ through a sparse dictionary of learned features. This design carries an implicit assumption that all meaningful content in the activation can be decomposed into sparse, approximately monosemantic directions, an assumption that emerged from how the field has operationalized the Toy Model of Superposition~\citep{Elhage2022} in practice.
 
Despite widespread adoption of this paradigm, multiple independent lines of evidence suggest the assumption may be too broad. \citet{Sun2025a} document persistent high fire rate latents in well-trained SAEs, classifying them into six functional categories (position tracking, context binding, nullspace, alphabet, part-of-speech, and PCA reconstruction). \citet{Engels2025a} identify a linearly predictable component in SAE residuals that standard architectures systematically fail to capture. Canonical massive-activation outlier dimensions~\citep{Bondarenko2023} resist sparse encoding and are known to concentrate in fewer than 20 channels out of thousands. These phenomena have been studied independently as separate pathologies of SAE training.
 
We hypothesize that these failure modes may reflect a shared structural source, namely activation content whose computational role is incompatible with sparse dictionary coding at standard scale. If such content exists, forcing it through a sparse dictionary should waste dictionary capacity on features that are neither interpretable nor monosemantic, while a dedicated low-dimensional channel should be able to absorb it, freeing the SAE to specialize on content that is amenable to sparse decomposition. This hypothesis makes a testable architectural prediction. A small parallel channel, if appropriately constrained, should simultaneously improve SAE quality metrics \textbf{and} reduce the prevalence of dense latents.
 
To test this prediction, we introduce a \textbf{rank-$r$ linear bottleneck} running parallel to standard SAEs. The SAE operates on the residual $\mathbf{x} - \text{sg}[\hat{\mathbf{x}}_\text{dense}]$, receiving only what the bottleneck does not capture. Functional separation is maintained by three design constraints, namely linearity (preventing the bottleneck from re-encoding sparse features), low rank (preventing SAE starvation), and gradient isolation (preventing co-adaptation). The bottleneck adds negligible parameters ($2 \times r \times D$, less than 0.7\% of an SAE with a 16{,}384-element dictionary) without requiring changes to the SAE's architecture or training procedure.
 
Our contributions are as follows:
\begin{itemize}
  \item We propose a simple and effective method for separating activation content that is unsuitable for sparse decomposition. By placing a low-rank linear bottleneck parallel to a standard SAE, we allow dense, low-dimensional structure to be absorbed before sparse reconstruction, without modifying the SAE's architecture or training procedure. On Gemma-2-2B layer~12, a rank-24 bottleneck reduces dense latent count by up to 84\% and improves sparse probing and targeted probe perturbation on both architectures at matched sparsity.
 
  \item Through causal intervention and post-hoc analysis, we establish that the absorbed component is computationally necessary (removing it raises cross-entropy 7.5$\times$, far exceeding the 2.8$\times$ from removing the geometrically near-identical top-24 PCA directions) and structurally identifiable as the top principal components and Bondarenko outlier dimensions. Sparse dictionaries encode this content redundantly rather than efficiently, with ablating 787 maximally aligned sparse features raising cross-entropy by only 2.9$\times$ and ablating 2,048 topic-aligned features leaving MMLU topic classification virtually unchanged, whereas removing the component drops it from 98.7\% to chance.
 
  \item We provide empirical evidence that the standard practice of training SAEs to reconstruct the entire residual stream is suboptimal. A subset of activation content is functionally important for downstream computation yet inherently unsuitable for sparse monosemantic decomposition, which we term a \textbf{computational scaffold}. Failing to separate it wastes dictionary capacity, produces persistent dense latents~\citep{Sun2025a}, and degrades downstream extraction quality. Our results suggest that sparsity-based interpretability methods may have a narrower scope of applicability than the field's current practice assumes.
\end{itemize}

%==============================================================================
\section{Related work}
\label{sec:related}

\paragraph{Sparse autoencoder architectures.}
Residual-stream SAEs were established as the standard interpretability tool by \citet{bricken2023monosemanticity, Cunningham2023a} and scaled to frontier models by \citet{templeton2024scaling}. Subsequent work has largely targeted the activation function used to enforce sparsity, including TopK~\citep{Gao2024}, JumpReLU~\citep{Rajamanoharan2024a}, Gated~\citep{Rajamanoharan2024}, BatchTopK~\citep{Bussmann2024}, Matryoshka nested dictionaries~\citep{Bussmann2025}, and mixture-of-experts routing~\citep{Mudide2025}, together with transcoder variants~\citep{Dunefsky2024a, Paulo2025a}, end-to-end objectives~\citep{Braun2024}, and large open suites~\citep{Lieberum2024b, He2024b} now standardly evaluated by~\citet{Karvonen2025}. Our contribution is orthogonal: we leave the SAE's architecture and training procedure untouched and add a small parallel linear bottleneck. Whereas \citet{Paulo2025a} use a full-rank affine skip on MLP transcoders, we apply a \textbf{low-rank} linear projection to \textbf{residual-stream SAEs} and show that the rank constraint is necessary to avoid SAE starvation (Appendix~\ref{sec:controls}).

\paragraph{Failure modes of sparse autoencoders.}
SAEs exhibit systematic pathologies that suggest structural limits to the sparse-decomposition assumption. \citet{Sun2025a} catalogue six classes of dense latents that persist in well-trained SAEs (position, context, nullspace, alphabet, part-of-speech, PCA reconstruction), and \citet{Engels2025a} show that the SAE error vector contains a ``dark matter'' component largely linearly predictable from the input, both indicating that a portion of activation content systematically resists sparse decomposition. Complementary failure modes include feature absorption and splitting~\citep{Chanin2025}, multi-dimensional non-linear features~\citep{Engels2025}, non-canonical and non-atomic feature units~\citep{Leask2025}, and seed-dependent decompositions~\citep{Paulo2025b}. On the evaluation side, SAEs are matched or beaten by simple baselines for probing~\citep{Kantamneni2025} and steering~\citep{Wu2025b}, are not distinguished from randomly initialized transformers by standard interpretability metrics~\citep{Heap2026}, and exhibit systematic gaps between proxy and downstream metrics~\citep{Karvonen2025}. We hypothesise that several of these phenomena, including persistent dense latents, dark matter, and the broader resistance of certain content to sparse coding, share a common structural source, and provide direct evidence in Section~\ref{sec:analysis}.

\paragraph{Structure of language model activations.}
Transformer residual streams exhibit pronounced structure beyond the feature-superposition picture~\citep{Elhage2022} that motivates SAE design. A long line of work documents extreme-magnitude outlier dimensions concentrated in a handful of channels~\citep{Kovaleva2021, Dettmers2022, Bondarenko2023, Sun2024}, closely tied to the attention-sink phenomenon~\citep{Xiao2024, Cancedda2024}, while the embedding-geometry literature shows that contextual representations are highly anisotropic and dominated by a few top principal components~\citep{Ethayarajh2019, Gao2019, Mu2018}. Section~\ref{sec:analysis_alignment} shows that these outlier and top-PC subspaces overlap substantially with the component our bottleneck learns to absorb, connecting the outlier-dimension and dense-latent literatures and suggesting both reflect the same low-rank activation structure.

%==============================================================================
\section{Method}
\label{sec:method}
%==============================================================================

\begin{figure}[htbp]
  \centering
  \includegraphics[width=0.85\textwidth]{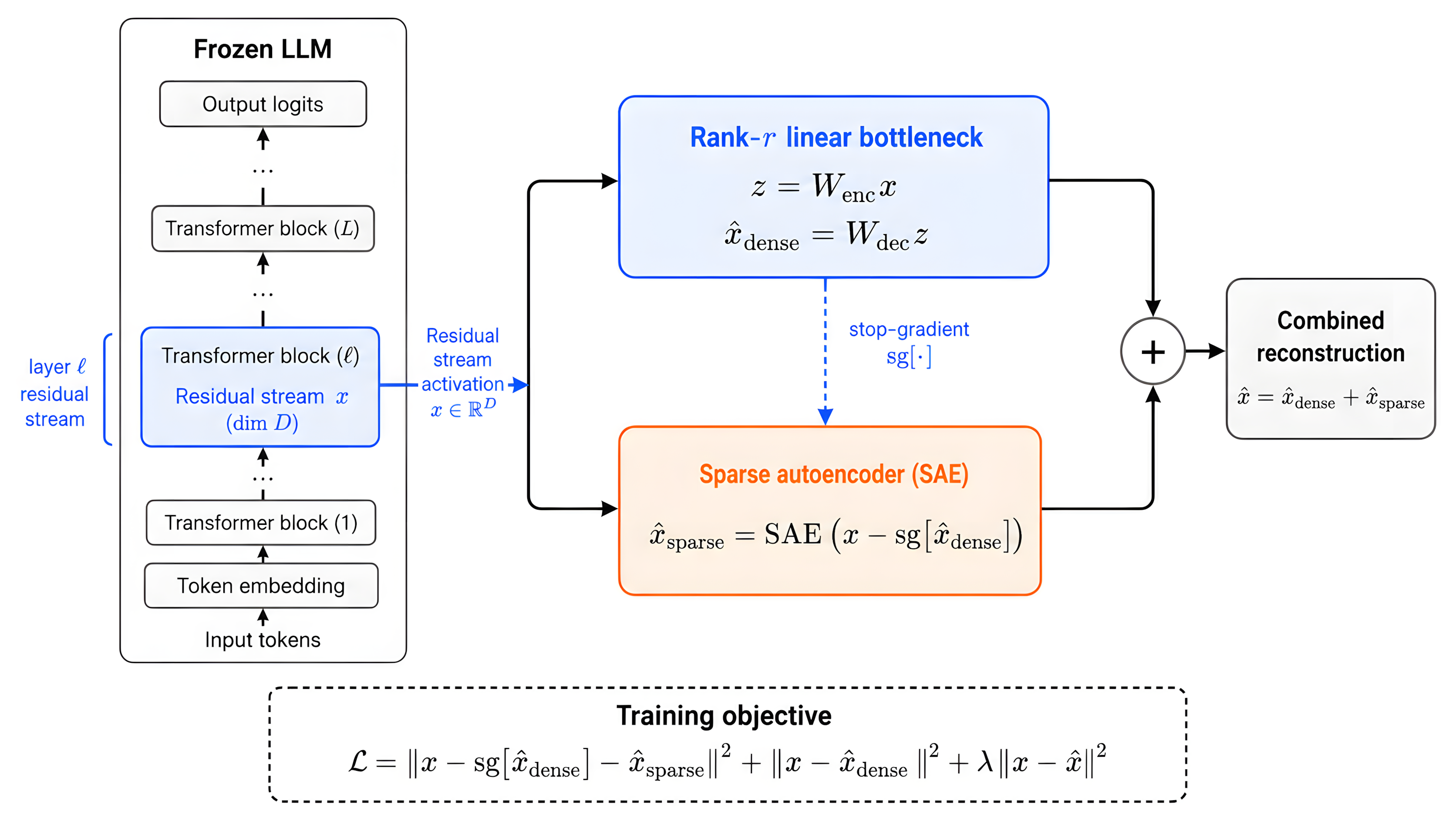}
  \caption{Architecture overview. A rank-$r$ linear bottleneck runs parallel to a standard sparse autoencoder. The SAE operates on the residual $\mathbf{x} - \text{sg}[\hat{\mathbf{x}}_\text{dense}]$ after the bottleneck's contribution is subtracted with stop-gradient. The combined reconstruction $\hat{\mathbf{x}} = \hat{\mathbf{x}}_\text{dense} + \hat{\mathbf{x}}_\text{sparse}$ is trained with a three-term loss (Eq.~\ref{eq:loss}).}
  \label{fig:architecture}
  \vspace{-2em}
\end{figure}

We place a linear bottleneck of rank $r$ parallel to a standard sparse autoencoder (Figure~\ref{fig:architecture}). Given activation $\mathbf{x} \in \RR^D$:
\begin{align}
  \mathbf{z} &= W_{\text{enc}} \mathbf{x}, \quad W_{\text{enc}} \in \RR^{r \times D} \label{eq:encode} \\
  \hat{\mathbf{x}}_{\text{dense}} &= W_{\text{dec}} \mathbf{z}, \quad W_{\text{dec}} \in \RR^{D \times r} \label{eq:decode} \\
  \hat{\mathbf{x}}_{\text{sparse}} &= \text{SAE}\bigl(\mathbf{x} - \text{sg}[\hat{\mathbf{x}}_{\text{dense}}]\bigr) \label{eq:sparse} \\
  \hat{\mathbf{x}} &= \hat{\mathbf{x}}_{\text{dense}} + \hat{\mathbf{x}}_{\text{sparse}} \label{eq:recon}
\end{align}
where $\text{sg}[\cdot]$ denotes stop-gradient (detach). The SAE receives only the residual after the bottleneck's contribution, and the bottleneck does not receive gradients through the SAE's loss. This architecture is compatible with any sparse SAE; we validate on BatchTopK ($|\mathcal{D}|{=}16384$, $k{=}40$) and MatryoshkaBatchTopK (with the same dictionary size and 5 nested groups).

The training loss combines three terms:
\begin{equation}
  \mathcal{L} = \underbrace{\|\mathbf{x}_{\text{sparse\_in}} - \hat{\mathbf{x}}_{\text{sparse}}\|^2}_{\text{sparse reconstruction}} + \underbrace{\|\mathbf{x} - \hat{\mathbf{x}}_{\text{dense}}\|^2}_{\text{bottleneck reconstruction}} + \lambda \underbrace{\|\mathbf{x} - \hat{\mathbf{x}}\|^2}_{\text{full-signal}} \label{eq:loss}
\end{equation}
with $\lambda{=}1.0$ and $\mathbf{x}_\text{sparse\_in} = \mathbf{x} - \text{sg}[\hat{\mathbf{x}}_\text{dense}]$. The role of each term, the architectural constraints (linearity, low rank, gradient isolation), the rank choice ($r{=}24$), and control experiments validating the linearity and rank constraints are detailed in Appendix~\ref{sec:controls}.

%==============================================================================
\section{Experiments}
\label{sec:experiments}
%==============================================================================

\subsection{Linear bottleneck improves SAE quality}
\label{sec:lift}

We first ask whether the bottleneck improves SAE quality. We test this by training BatchTopK and Matryoshka SAEs with and without a rank-24 bottleneck on Gemma-2-2B (layer~12) at $L_0{=}40$, and evaluating with the SAEBench full suite~\citep{Bussmann2025} (training details in Appendix~\ref{app:training}). Table~\ref{tab:lift} presents the effect of adding a rank-24 linear bottleneck to two SAE architectures under otherwise identical training conditions. On both architectures, extraction quality metrics improve consistently. Explained variance increases by 0.7--1.2 percentage points, sparse probing top-1 accuracy improves by 0.9--1.4 points, and targeted probe perturbation (TPP) at threshold 20 roughly doubles. The number of dead features also decreases on both architectures, suggesting that absorbing dense content frees dictionary capacity.

While regressions occur on intervention-based metrics (SCR, RAVEL) and autointerp on BatchTopK, these are architecture-dependent and do not appear on Matryoshka. We analyze these effects in Appendix~\ref{app:scr} and~\ref{app:autointerp}.

\begin{table}[t]
  \caption{Effect of adding a rank-24 linear bottleneck to BatchTopK and Matryoshka SAEs. All metrics at strict $L_0{=}40$, matched training conditions. $\uparrow$/$\downarrow$ indicate preferred direction. SP, Absorption, and Autointerp are averaged over 3 random seeds. SCR reports direction-1 (spurious-attribute ablation; see Appendix~\ref{app:scr} for metric definition). The bottleneck improves extraction quality metrics (EV, SP, TPP) on both architectures, while intervention based metrics (SCR, RAVEL) show architecture dependent effects analyzed in Appendix~\ref{app:scr}. Autointerp is unchanged on Matryoshka and shows a small ($-$1.2 pp) regression on BatchTopK that is distributed across fire-rate bins (Appendix~\ref{app:autointerp}).}
  \label{tab:lift}
  \centering
  \small
  \begin{tabular}{lccclccc}
    \toprule
    & \multicolumn{3}{c}{\textbf{BatchTopK}} & & \multicolumn{3}{c}{\textbf{Matryoshka}} \\
    \cmidrule(lr){2-4} \cmidrule(lr){6-8}
    Metric & $r{=}0$ & $r{=}24$ & $\Delta$ & & $r{=}0$ & $r{=}24$ & $\Delta$ \\
    \midrule
    EV $\uparrow$ & .793 & \textbf{.805} & +.012 & & .813 & \textbf{.820} & +.008 \\
    SP top-1 $\uparrow$ & .749 & \textbf{.758} & +.009 & & .765 & \textbf{.779} & +.014 \\
    SP top-5 $\uparrow$ & .851 & \textbf{.864} & +.013 & & .881 & \textbf{.886} & +.005 \\
    TPP@20 $\uparrow$ & .023 & \textbf{.054} & +.031 & & .063 & \textbf{.087} & +.024 \\
    Absorption $\downarrow$ & \textbf{.057} & .084 & +.027 & & .011 & \textbf{.009} & $-$.002 \\
    SCR@10 $\uparrow$ & \textbf{+.213} & $-$.295 & $-$.508 & & +.263 & \textbf{+.303} & +.040 \\
    RAVEL cause $\uparrow$ & \textbf{.582} & .558 & $-$.024 & & .570 & \textbf{.592} & +.022 \\
    Autointerp $\uparrow$ & \textbf{.869} & .857 & $-$.012 & & .865 & .865 & tied \\
    \midrule
    $n_\text{dense}$ $\downarrow$ & 25 & \textbf{4} & $-$21 & & 69 & \textbf{44} & $-$25 \\
    $n_\text{dead}$ $\downarrow$ & 439 & \textbf{305} & $-$134 & & 1369 & \textbf{1229} & $-$140 \\
    \bottomrule
  \end{tabular}
  \vspace{-2em}
\end{table}

\begin{figure}[t]
  \centering
  \includegraphics[width=0.9\textwidth]{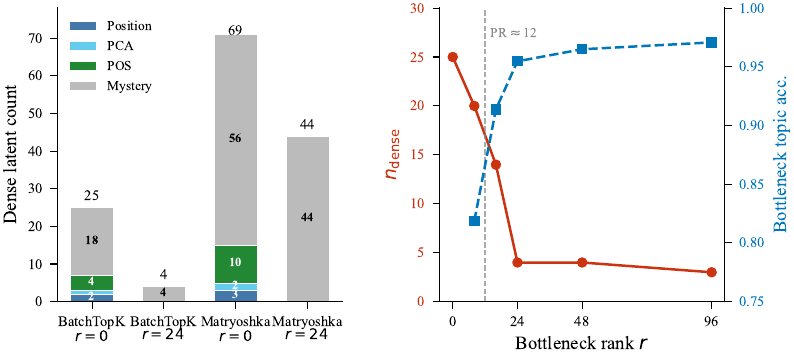}
  \caption{The bottleneck eliminates classifiable dense latents and saturates near the activation's natural rank. \textbf{Left}: dense latent count and category breakdown at $r{=}0$ vs.\ $r{=}24$ on both architectures. Classifiable categories (position, PCA, POS) vanish by $r{=}24$; residual dense latents are unclassified. \textbf{Right}: BatchTopK rank sweep ($r \in \{0, 8, 16, 24, 48, 96\}$). $n_\text{dense}$ (left axis, red) drops sharply at $r{=}24$ and saturates; MMLU topic accuracy from the bottleneck latent (right axis, blue) increases monotonically and plateaus beyond $r{=}24$. Dashed line marks the participation ratio (PR$\approx$12). Full numerical values in Appendix~\ref{app:rank_sweep}.}
  \label{fig:dense_and_rank}
\vspace{-2em}
\end{figure}

\subsection{Absorbing the component reduces dense latent count}
\label{sec:analysis_dense}

The improvements in Section~\ref{sec:lift} suggest that the bottleneck absorbs content that would otherwise burden the sparse dictionary. If our hypothesis is correct, this content should overlap with the persistent dense latents documented by \citet{Sun2025a}. Indeed, the dense latent count (features with fire rate $> 0.1$) drops from 25 to 4 at $r{=}24$ on BatchTopK and from 69 to 44 on Matryoshka, saturating at higher ranks (Figure~\ref{fig:dense_and_rank}, left).

Using the dense latent categories defined by \citet{Sun2025a}, we find that the bottleneck eliminates all classifiable dense latents (position tracking, PCA-aligned, and part-of-speech categories) on both architectures, with only unclassified dense latents remaining at $r{=}24$ (Figure~\ref{fig:dense_and_rank}, left). 

\subsection{Sweeping bottleneck capacity confirms low-rank structure}
\label{sec:sweep}
We further verify whether the absorbed component is low-rank by training a family of BatchTopK SAEs across $r \in \{0, 8, 16, 24, 48, 96, 2304\}$ with all other hyperparameters held fixed. Figure~\ref{fig:dense_and_rank} (right) presents two key metrics across this sweep.

Dense latent count ($n_\text{dense}$, features with fire rate $> 0.1$) drops sharply at $r{=}24$ and saturates at higher ranks. Topic classification accuracy from the bottleneck latent (a linear probe trained on the $r$-dimensional representation to predict MMLU subject) increases monotonically, reaching 95\% at $r{=}24$ with marginal gains beyond. This empirical saturation aligns with the structural rank estimate of ${\sim}12$ effective dimensions (Appendix~\ref{sec:controls}), so that $r{=}24$ provides approximately twice the estimated natural rank and both metrics confirm the absorbed component is low-rank.

The $r{=}2304$ endpoint serves as a starvation control. When the bottleneck has full-rank capacity, it absorbs the entire activation (EV=1.00) and the SAE degenerates to noise-level feature selection (SP top-1 = .661). This confirms that the low-rank constraint is a necessary design element.

%==============================================================================
\section{Characterizing the absorbed content as a computational scaffold}
\label{sec:analysis}
%==============================================================================
We now characterize the absorbed content and argue that it constitutes a \textbf{computational scaffold}, a component that carries coarse contextual information, that the model's downstream computation depends on, and that resists efficient sparse encoding. We establish this through the scaffold's structural identity (Section~\ref{sec:analysis_alignment}), causal necessity (Section~\ref{sec:causal}), and redundant representation under sparse dictionaries (Section~\ref{sec:analysis_topic}).

\subsection{The scaffold aligns with top principal components and outlier dimensions}
\label{sec:analysis_alignment}

\begin{wrapfigure}{r}{0.45\textwidth}
  \centering
  \includegraphics[width=0.43\textwidth]{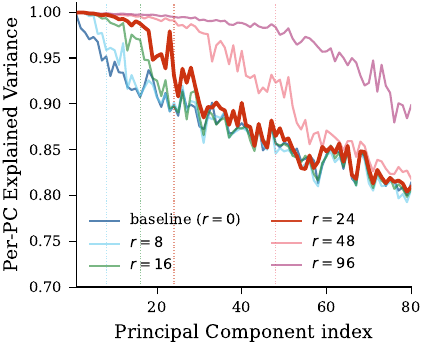}
  \caption{Per-PC explained variance as a function of principal component index. }
  \label{fig:per_pc_ev}
   \vspace{-10pt}
\end{wrapfigure}

To characterize what the bottleneck learns, we measure how much of each principal component's variance is captured by its reconstruction $\hat{\mathbf{x}}_\text{dense}$. Figure~\ref{fig:per_pc_ev} shows a clean step-function pattern, where a rank-$r$ bottleneck explains $>$99\% of the variance along each of the first ${\sim}r$ principal components, then drops sharply to the $r{=}0$ baseline. The plateau breakpoint tracks the rank, confirming that the bottleneck converges to approximately the top-$r$ PCA subspace, as expected from minimizing rank-$r$ reconstruction MSE.

\begin{figure}[t]
  \centering
  \includegraphics[width=0.9\textwidth]{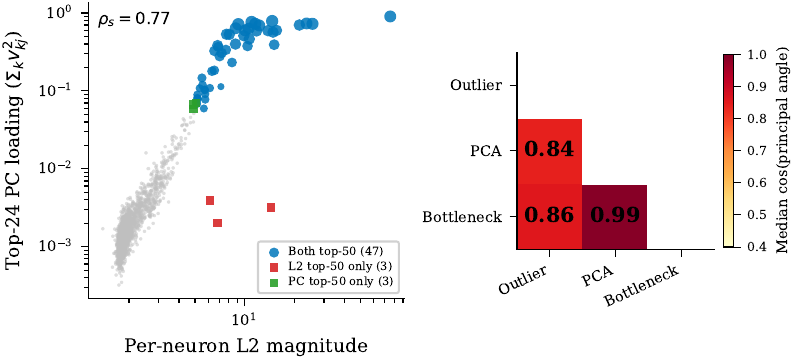}
  \caption{Alignment between outlier dimensions, top principal components, and the learned bottleneck. \textbf{Left:} Per-neuron L2 magnitude vs.\ top-24 PC loading for all 2304 residual-stream neurons (log--log). Blue circles mark the 47 neurons in both the L2-magnitude and PC-loading top-50; dot size is proportional to the bottleneck encoder weight $\|W_{\mathrm{enc}}[:,j]\|$. Red and green squares mark the 3 neurons in only one top-50 set. \textbf{Right:} Median cosine of principal angles between three 24-dimensional subspace definitions. The bottleneck nearly coincides with the PCA subspace (0.99) and aligns strongly with outlier axes (0.86).}
  \label{fig:alignment}
  \vspace{-1em}
\end{figure}

We next ask whether this top-variance subspace corresponds to the outlier dimensions previously documented in transformer activations. Per-neuron L2-magnitude rank and top-24 PC loading rank correlate at Spearman $\rho = 0.77$, with 47 of the top 50 neurons by each measure coinciding (Figure~\ref{fig:alignment}, left). The dominant-variance directions that the bottleneck absorbs thus largely coincide with the canonical outlier dimensions identified by \citet{Bondarenko2023}.

Finally, we compare the three definitions of the 24-dimensional component directly. The subspaces defined by outlier axes, top PCA directions, and the learned bottleneck yield pairwise median cosine similarities between 0.84 and 0.99 (Figure~\ref{fig:alignment}, right). Three operationally distinct procedures, namely magnitude ranking, variance decomposition, and end-to-end learning, converge on the same structural component of the residual stream.

\subsection{Removing the scaffold disrupts next-token prediction}
\label{sec:causal}

\begin{figure}[t]
  \centering
  \includegraphics[width=\textwidth]{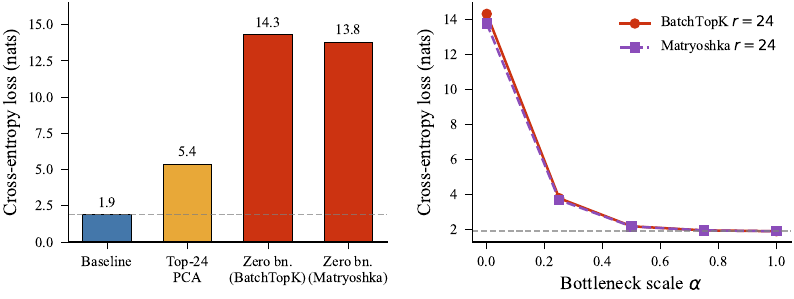}
  \caption{Causal intervention on the bottleneck's learned 24-d component. \textbf{Left}: condition comparison. Removing the bottleneck component (zero) raises CE significantly on both architectures, removing a random 24-d component has negligible effect, and top-24 PCA removal is intermediate. \textbf{Right}: dose-response. Scaling the bottleneck contribution $\alpha \in [0.25, 1.0]$ produces a smooth CE gradient. Dashed lines mark baseline and zero bottleneck reference.}
  \label{fig:intervention}
  \vspace{-2em}
\end{figure}

To test the causal importance of this component, we project out the bottleneck's learned 24-dimensional subspace from layer-12 activations and measure the effect on next-token prediction (full protocol in Appendix~\ref{app:causal_protocol}). Against a baseline of CE $= 1.90$, removing the component raises CE to 14.3, a 7.5$\times$ increase. Two independently trained bottlenecks (BatchTopK and Matryoshka, both $r{=}24$) produce nearly identical effects (CE $14.3$ and $13.8$), confirming that the learned component is architecturally stable. Scaling the bottleneck contribution by $\alpha \in [0, 1]$ yields a smooth monotonic dose-response (Figure~\ref{fig:intervention}, right), ruling out a binary on/off role and indicating that the model depends continuously on this component's magnitude.

Notably, despite the close geometric alignment between the bottleneck and the top principal components established in Section~\ref{sec:analysis_alignment}, removing the top-24 PCA directions raises CE to only 5.37 (Figure~\ref{fig:intervention}, left). This large gap in causal effect suggests that joint training with the SAE steers the bottleneck toward computationally critical directions within the top-PC neighborhood that static PCA does not isolate.

\subsection{Sparse encoding of the scaffold is redundant}
\label{sec:analysis_topic}

The results above establish that the scaffold is causally important and geometrically identifiable. We now ask whether the sparse dictionary encodes this content efficiently or redundantly. If the SAE had concentrated the scaffold into a small number of dedicated features, ablating them should approximate the effect of removing the component directly. If instead the scaffold is spread thinly across many features that each partially track it, no tractable subset should suffice.

\begin{figure}[t]
  \centering
  \includegraphics[width=0.9\textwidth]{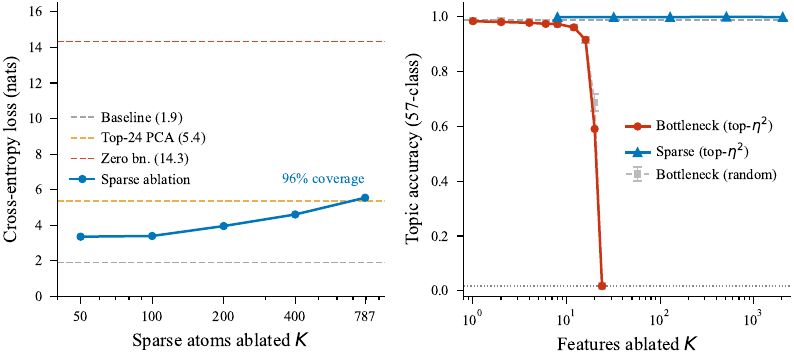}
  \caption{Sparse encoding of the scaffold is redundant. \textbf{Left}: CE loss vs.\ top-$K$ sparse features ablated by decoder $\alpha$-alignment with the bottleneck column space. \textbf{Right}: topic accuracy (57-class MMLU) under progressive ablation. Bottleneck dim ablation produces a sharp cliff at $K{\approx}20$, with random ordering tracking top-$\eta^2$ closely (topic is distributed uniformly across the 24 dims). }
  \label{fig:sparse_redundancy}
  \vspace{-2em}
\end{figure}

We rank all 16,384 SAE features by decoder weight alignment with the bottleneck column space and progressively ablate them. Even at $K{=}787$, where the ablated decoder vectors span 96\% of the scaffold subspace, CE rises to only 5.5 (Figure~\ref{fig:sparse_redundancy}, left), far short of the 14.3 produced by zeroing the 24-dimensional component itself. Despite near-complete geometric coverage, sparse ablation cannot replicate the causal effect, because these features are activation-gated and fire only on matching inputs, leaving substantial residual signal in the scaffold directions.

The same asymmetry holds for coarse topic information. A linear probe on the bottleneck's 24-dimensional latent achieves 95\% accuracy on MMLU subject classification (57 classes, binary OvR), approaching the full sparse code's ceiling of 98\%. The multinomial 57-way variant used for the ablation curves in Figure~\ref{fig:sparse_redundancy} (right) reports the same gap on a different scale (98.7\% vs.\ 99.8\%), yet both protocols agree that 24 bottleneck dimensions recover nearly all of the sparse code's topic information. Ablating bottleneck dimensions by descending $\eta^2$ produces a sharp cliff at $K{\approx}20$, reaching chance by $K{=}24$. By contrast, ablating the top 2,048 sparse atoms by $\eta^2$ leaves accuracy above 99\%. The same topic-level content is compactly encoded in 24 bottleneck dimensions but redundantly distributed across the sparse dictionary.

%==============================================================================
\section{Discussion: The elephant in the interpretability room}
\label{sec:discussion}
%==============================================================================

\subsection{The scope of sparsity-based interpretability.}
A common working assumption behind sparse-dictionary interpretability is that the full residual stream should be decomposed into sparse, approximately monosemantic features. Prior work has already identified various symptoms suggesting this assumption may be too broad, from persistent dense latents~\citep{Sun2025a} to linearly predictable reconstruction error~\citep{Engels2025a}, metric-validity concerns~\citep{Heap2026, Karvonen2025}, and feature non-atomicity~\citep{Leask2025}. These have largely been studied as separate pathologies. Our work suggests they may share a common structural source. The component we isolate occupies only ${\sim}24$ dimensions of the residual stream, yet it accounts for the dominant share of activation variance (Section~\ref{sec:analysis_alignment}), proves to be the most causally important structure we can identify (Section~\ref{sec:causal}), and carries semantically meaningful content (Section~\ref{sec:analysis_topic}). At the same time, it is the content that sparse dictionaries represent least efficiently, with the sparse code tracking it in a distributed and redundant fashion that fails to approximate its causal effect (Section~\ref{sec:analysis_topic}).
 
These observations motivate interpreting the bottleneck-absorbed component as a \textbf{computational scaffold}, activation content that is not naturally represented as a small set of sparse monosemantic features yet appears functionally important for downstream computation. This interpretation is supported by three converging lines of evidence, namely dominant variance share, severe degradation upon removal, and semantically structured information content. The top-PC/outlier component we isolate is one instance of this category; other forms of scaffold content with different geometric profiles may well exist in residual streams (see Open Questions below). The most causally important component of the residual stream appears to be the one that the field's primary interpretability tool handles least well. We hope this work serves as an invitation to confront the elephant directly.

\subsection{Implications for evaluation.}
Current evaluation frameworks for sparse features share an implicit premise, precisely that the properties being measured should be isolable to a tractable number of sparse features, which may not be the best way of encoding those properties in itself. SCR's ablation protocol, for instance, registers the scaffold's migration to the bottleneck as a failure of the SAE, while joint ablation of bottleneck and sparse components achieves cleaner disentanglement than sparse intervention alone (Appendix~\ref{app:scr}).
 
Autointerp exhibits a complementary blind spot. In our cases, autointerp score fails to reflect the desirable properties after the bottleneck absorbs persistent dense latents (Appendix~\ref{app:autointerp}). More broadly, \citet{Heap2026} demonstrate that autointerp scores fail to distinguish trained transformers from randomly initialized ones, suggesting that the metric is insensitive to computational relevance in the first place. Both observations suggest that these metrics, while useful, may need to be supplemented with evaluations that account for content allocated outside the sparse dictionary.
 
\subsection{Reinterpreting the success of Matryoshka SAEs.}
Matryoshka SAEs~\citep{Bussmann2025} were designed to address feature absorption through nested prefix dictionaries that learn multi-level semantic hierarchies. Our analysis suggests a complementary mechanism that may contribute to their success. Their first group $g_0$ carries a disproportionate share of the total explained variance and dense latent population relative to its size, functioning in practice as an implicit semi-dense channel for scaffold content (Appendix~\ref{app:matry_g0}). When an explicit bottleneck is introduced, it subsumes this role, offloading scaffold content from $g_0$ and freeing capacity for later groups to specialize. The nested prefix design appears to have created, perhaps inadvertently, a mechanism for scaffold absorption. The semantic-hierarchy interpretation and the scaffold-absorption mechanism are compatible, and may both contribute to Matryoshka's empirical advantage; the latter, however, has not been previously recognized.
 
\subsection{Open questions.}
We view our contribution as identifying the first explicit instance of a computational scaffold. This raises more questions than it answers, and we organize the most pressing ones along three directions.
 
\textbf{What else is scaffold?}
Our method isolates top-variance content by construction. Content that is distributed broadly across the principal component spectrum, such as sentiment (Appendix~\ref{app:sentiment}), falls outside the bottleneck's absorption scope. Whether scaffold-like content exists in the middle variance range, where our current method cannot reach, remains open. The residual dense latents that persist after bottleneck absorption, unclassified by any test in the taxonomy of \citet{Sun2025a}, hint at scaffold content beyond the top-variance regime. Similarly, the ``dark matter'' of \citet{Engels2025a} is not substantially reduced by our bottleneck, raising the possibility that it constitutes a distinct form of scaffold with a different geometric profile.

\textbf{What computational role does the scaffold play?} 
The stark causal asymmetry between removing the scaffold and ablating large numbers of scaffold-aligned sparse features (Section~\ref{sec:analysis_topic}) suggests that sparse features may depend on the scaffold as a substrate providing coarse contextual information, without which fine-grained semantic distinctions cannot be effectively utilized. We emphasize that this remains an interpretive hypothesis. The scaffold is superficially consistent with the Toy Model of Superposition~\citep{Elhage2022}, which predicts that high-frequency features occupy near-orthogonal directions, yet its dense, low-rank, collectively encoded character is unlike any feature the TMS framework considers. Whether it represents an extreme TMS regime or a fundamentally different representational structure is an important open question.
 
\textbf{What is the geometric structure of the scaffold?}
The learned bottleneck and the top PCA subspace are geometrically near-identical yet causally divergent (Section~\ref{sec:causal}), indicating that joint training with the SAE steers the bottleneck toward a causally privileged subset of directions within the top-PC neighborhood. This gap between geometric overlap and causal effect suggests that the computationally critical structure within the top-variance subspace is finer-grained than PCA alone can resolve. Understanding the origin of this discrepancy may shed light on how transformers organize functionally distinct information within shared variance-dominant subspaces.
 
\subsection{Limitations.}
Our experiments use a single model and layer (Gemma-2-2B, layer 12). Because our central claim is that the working assumption of full sparse decomposability might be too broad, one well-characterized counterexample suffices to raise the question; nevertheless, cross-model and cross-layer replication would strengthen confidence in the generality of the architectural prescription. Rank selection ($r{=}24$) is justified empirically for this setting, and whether the optimal rank scales with model size or layer depth remains an open question.
 
Our analysis is restricted to residual stream SAEs. Whether MLP transcoders~\citep{Dunefsky2024a, Paulo2025a} and attention-output SAEs face analogous scaffold phenomena is unknown. The residual stream aggregates information from all model components, making it a natural locus for low-rank broadcast structure; other decomposition targets may exhibit different activation geometry, and the relevance of our findings to those settings requires independent investigation.

%==============================================================================
\section{Conclusion}
\label{sec:conclusion}
%==============================================================================
 
We have shown that language model residual streams contain a compact, low-rank component that is causally critical for downstream computation yet poorly suited to sparse monosemantic decomposition (a computational scaffold). A simple rank-$r$ linear bottleneck running parallel to a standard SAE suffices to absorb this component, reducing dense latent counts by up to 84\% and improving extraction quality metrics on two SAE architectures without modifying the SAE itself. The absorbed component aligns with the top principal components and outlier dimensions, carries coarse semantic content, and is encoded redundantly rather than efficiently by sparse dictionaries. These findings suggest that the common practice of training SAEs to reconstruct the entire residual stream conflates two structurally distinct kinds of activation content, and that explicitly separating them yields better sparse decompositions. More broadly, our results indicate that the scope of sparsity-based interpretability warrants re-examination, as the most causally important structure in the residual stream may be precisely what current methods handle least well.

%==============================================================================
\begin{ack}
% To be filled after acceptance.
\end{ack}
%==============================================================================

\newpage
\bibliographystyle{abbrvnat}
\bibliography{references}

%%%%%%%%%%%%%%%%%%%%%%%%%%%%%%%%%%%%%%%%%%%%%%%%%%%%%%%%%%%%
\appendix

\section{Matryoshka $g_0$ analysis}
\label{app:matry_g0}

\begin{figure}[h]
  \centering
  \includegraphics[width=0.65\textwidth]{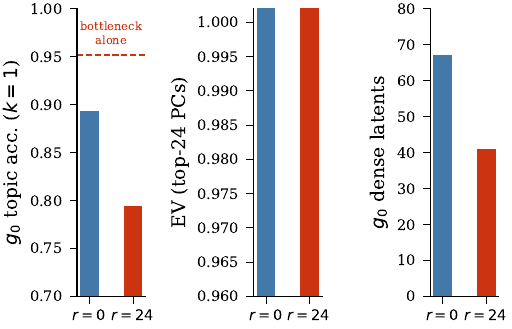}
  \caption{Matryoshka $g_0$ before and after adding the bottleneck. \textbf{Left}: $g_0$ topic accuracy ($k{=}1$) drops 10 pp when the bottleneck absorbs topic content (dashed line shows the bottleneck alone achieves 95\%). \textbf{Center}: per-PC EV on top-24 PCs improves. \textbf{Right}: $g_0$ dense latent count drops from 67 to 41. The explicit bottleneck subsumes the implicit scaffold absorption previously performed by $g_0$.}
  \label{fig:matry_g0}
\end{figure}

Matryoshka SAEs~\citep{Bussmann2025} use nested prefix dictionaries to address feature absorption. We observe that their first group $g_0$ (512 atoms) functions as an implicit semi-dense channel, carrying 78\% of total EV, harboring 67 of 69 dense latents, and achieving 89.3\% topic accuracy at $k{=}1$. Adding the explicit bottleneck subsumes this role (Figure~\ref{fig:matry_g0}): $g_0$'s topic accuracy drops 10 pp (while the bottleneck alone reaches 95\%), its dense latent count drops from 67 to 41, and freed capacity allows later groups to specialize more effectively.

\section{Design choices and validation}
\label{sec:controls}

\paragraph{Three-term loss.} The three terms in Eq.~\ref{eq:loss} serve complementary roles. The sparse reconstruction loss trains the SAE to reconstruct the bottleneck's residual $\mathbf{x}_\text{sparse\_in} = \mathbf{x} - \text{sg}[\hat{\mathbf{x}}_\text{dense}]$, and the stop-gradient ensures this loss does not send gradients to the bottleneck. The bottleneck reconstruction loss provides the bottleneck with an independent training signal, driving it toward the minimum-MSE rank-$r$ projection. This term is mandatory, as removing it deprives the bottleneck of gradient signal and causes degeneration. The full-signal loss ($\lambda{=}1.0$) provides a joint training signal to both components, encouraging coherent combined reconstruction of $\mathbf{x}$. This design differs from naive joint training ($\|\mathbf{x} - \hat{\mathbf{x}}\|^2$ alone): under pure joint training, the bottleneck would receive gradients only for filling residual gaps left by the SAE, potentially converging to an arbitrary or degenerate rank-$r$ complement. The standalone bottleneck loss ensures the bottleneck converges toward the top-$r$ principal components regardless of the SAE's behavior, providing a stable functional separation.

The four architectural constraints follow, with control experiments validating the linearity and rank constraints inline.

\paragraph{Linearity.} A nonlinear bottleneck (GELU activation) can re-encode features the SAE needs, defeating the functional separation. Empirically, a GELU-activated bottleneck at rank 24 yields EV = .793 (identical to the $r{=}0$ baseline), SP top-1 = .721 (below baseline), $n_\text{dense}=22$ (vs.\ 4 for linear), and bottleneck topic accuracy of 78\% (vs.\ 95\% for linear). Classification using the taxonomy of \citet{Sun2025a} confirms the failure, as the GELU variant retains 3 position tracking dense latents (vs.\ 0 for linear at the same rank), indicating that the nonlinear bottleneck does not absorb broadcast content. The likely mechanism is that GELU enables the bottleneck to re-encode sparse-feature-like content, competing with the SAE rather than complementing it. The fact that SP top-1 \textbf{decreases} below baseline with GELU (while remaining above baseline with linear) suggests active interference between the two components when nonlinearity is present.

\paragraph{Low rank.} A full-rank ($r{=}D{=}2304$) linear bottleneck absorbs the entire activation, starving the SAE to chance-level performance. The starvation control ($r{=}2304$, full-rank linear) yields EV = 1.00, leaving the SAE with negligible residual to decompose. Sparse probing drops to .661, RAVEL cause to .002, and the SAE's 40 active features become effectively noise-selected. This confirms that the rank constraint is what maintains the cooperative relationship between the two components.

\paragraph{Rank selection.} We set $r{=}24$ based on structural observations made \textbf{prior to training}. The participation ratio of the activation's per-axis magnitude is 12.3, indicating ${\sim}12$ effective dimensions for the heavy-variance structure. Three operationally distinct 24-dimensional subspaces (outlier axes, top PCA, and the learned bottleneck) align with pairwise median cosine similarity 0.84--0.99 (Section~\ref{sec:analysis_alignment}). Per-PC explained variance near-saturates at $r{=}24$.

\paragraph{Gradient isolation.} The stop-gradient in Eq.~\ref{eq:sparse} ensures the SAE optimizes reconstruction of $\mathbf{x} - \hat{\mathbf{x}}_\text{dense}$ without influencing the bottleneck's learned component. Without this isolation, the two components can co-adapt, undermining the functional separation.

Together, these constraints establish a design envelope: the bottleneck must be linear (to prevent feature re-encoding) and low-rank (to prevent starvation), with $r \ll D$ as a hard requirement.

\section{Sentiment direction analysis}
\label{app:sentiment}

To characterize the boundary of what the bottleneck absorbs, we computed the alignment between a CAA-style sentiment direction $\mathbf{v}_\text{sentiment}$ and the top-24 principal components of Gemma-2-2B L12. The sentiment direction requires 321 PCs to capture 50\% of its norm, and the top-24 PCs hold only 26.2\% of its energy. Accordingly, the bottleneck captures approximately 20\% of $\mathbf{v}_\text{sentiment}$. A linear probe on the bottleneck's 24 latent dimensions for sentiment achieves 60\% accuracy (vs.\ 71\% from sparse features alone). The bottleneck absorbs \textbf{top variance} content. Content that is distributed across many principal components falls outside its absorption scope regardless of semantic interpretability.

\section{Rank sweep full results}
\label{app:rank_sweep}

Table~\ref{tab:rank_sweep_full} provides the complete numerical results for the BatchTopK rank sweep summarized in Figure~\ref{fig:dense_and_rank} (right).

\begin{table}[h]
  \caption{Full rank sweep on BatchTopK. Topic = MMLU 57-class accuracy from bottleneck latent only. The $r{=}2304$ row is the full-rank starvation control.}
  \label{tab:rank_sweep_full}
  \centering
  \small
  \begin{tabular}{rcccccccc}
    \toprule
    Rank & EV & SP top-1 & TPP@20 & $n_\text{dense}$ & Topic & EV$_{\text{top24}}$ & EV$_{\text{tail500+}}$ & FVU$_\text{nl}$ \\
    \midrule
    0    & .793 & .749 & .023 & 25 & ---  & .961 & .311 & .106 \\
    8    & .797 & .759 & .081 & 20 & .82  & .973 & .315 & .104 \\
    16   & .797 & .742 & .111 & 14 & .91  & .985 & .317 & .102 \\
    24   & .805 & .761 & .054 & \textbf{4}  & .95  & .993 & .330 & .100 \\
    48   & .805 & .771 & .063 & 4  & .96  & .998 & .321 & .096 \\
    96   & .816 & .768 & .038 & 3  & .97  & .999 & .327 & .090 \\
    2304 & 1.00 & .661 & .062 & 45 & ---  & 1.00 & 1.00 & .000 \\
    \bottomrule
  \end{tabular}
\end{table}

\section{Training details}
\label{app:training}

We train on Gemma-2-2B (base) at layer~12 (\texttt{resid\_post}) with a dictionary of 16{,}384 elements, effective $L_0{=}40$ via BatchTopK with strict top-$k$ enforcement at evaluation, using \texttt{monology/pile-uncopyrighted} for approximately 500M tokens at context length 1024 with seed~0. The bottleneck adds $2rD$ parameters ($2 \times 24 \times 2304 = 110{,}592$ for $r{=}24$, or 0.67\% of the SAE's parameters).

% TODO: Full hyperparameter tables, learning rate schedules, etc.

\section{Architecture-dependent SCR effects}
\label{app:scr}

\paragraph{Metric definition.} SAEBench's SCR evaluation computes two directional scores per probe pair. Direction~1 ablates features encoding the spurious attribute (e.g., gender) and measures whether the target attribute (e.g., profession) prediction recovers toward the unbiased baseline. Direction~2 performs the reverse ablation. The default \texttt{scr\_metric} field selects whichever direction has lower clean accuracy, which in practice equals direction~2 for most probe pairs in our evaluation. We report direction~1 in Table~\ref{tab:lift} because it directly measures the intended SCR objective (removing spurious-attribute features to improve target prediction). Under this definition, the regression on BatchTopK is stark ($+.213 \to -.295$), while the default \texttt{scr\_metric} would show a mild positive change ($+.239 \to +.221$) that masks the direction-1 collapse.

The same rank-24 linear bottleneck produces \textbf{opposite} effects on SCR direction-1 depending on the host architecture. On BatchTopK, SCR drops from +.213 to $-$.295, while on Matryoshka it improves from +.263 to +.303.

Disaggregating by probe family reveals the source. On BatchTopK, the regression concentrates in the \texttt{bias\_in\_bios} probe (gendered-profession shortcut), where SCR direction-1 improves by +14 to +23 percentage points at thresholds 10--50 while SCR metric (averaged over both directions) degrades. The \texttt{amazon\_reviews} probe (sentiment $\times$ category) remains positive across ranks. The bottleneck absorbs low-rank content that BatchTopK's sparse features were incidentally capturing alongside sparse structure. When this content moves to the bottleneck, sparse feature intervention loses access to it, and the SCR metric evaluates this as regression.

To understand this dissociation mechanistically, we conduct a dual-attribute ablation on Matryoshka $r{=}24$ using two SCR probe families (Table~\ref{tab:scr_mechanism}). For \texttt{bias\_in\_bios} (gender $\times$ profession), the bottleneck does not absorb the spurious attribute itself ($\eta^2_\text{max} = 0.012$). Instead, the bottleneck's presence restructures the SAE, concentrating gender information into a small number of atoms (max $\eta^2 = 0.120$, versus 0.020 in the baseline). The small-$K$ ablation protocol then hits these concentrated atoms efficiently, producing apparent regression. For \texttt{amazon\_reviews} (sentiment $\times$ category), the bottleneck partially absorbs sentiment ($\eta^2_\text{max} = 0.174$) and simultaneously spreads the residual encoding more diffusely across the SAE (max atom $\eta^2$ drops from 0.256 to 0.160). The ablation finds insufficient concentration to exploit, and the score does not regress.

Joint ablation (zeroing the top-8 bottleneck dimensions and top-128 sparse atoms ranked by $\eta^2$) drives the gender probe to chance (0.503) while preserving profession accuracy at 0.801, demonstrating that the combined bottleneck+SAE system achieves cleaner disentanglement than either component alone when both are intervened upon jointly.

\begin{table}[h]
  \caption{Dual-attribute ablation on Matryoshka $r{=}24$ for two SCR probe families. The bottleneck restructures the SAE differently depending on the spurious attribute's rank profile. Gender (low-rank) concentrates in sparse atoms, while sentiment (higher-rank) is partially absorbed by the bottleneck and diffused across the sparse code.}
  \label{tab:scr_mechanism}
  \centering
  \small
  \begin{tabular}{lcc}
    \toprule
    Quantity & \texttt{bias\_in\_bios} & \texttt{amazon\_reviews} \\
    \midrule
    Bottleneck $\eta^2$(spurious) max dim & 0.012 & 0.174 \\
    Sparse $\eta^2$(spurious) max ($r{=}0$) & 0.020 & 0.256 \\
    Sparse $\eta^2$(spurious) max ($r{=}24$) & \textbf{0.120} ($6\times$) & 0.160 ($0.6\times$) \\
    Bottleneck restructures sparse & concentrates & spreads \\
    Joint ablation, target acc. & 0.801 & 0.417 \\
    Joint ablation, spurious acc. & \textbf{0.503} (chance) & 0.778 \\
    \bottomrule
  \end{tabular}
\end{table}

This pattern has implications for the SCR metric. Its small-$K$ ablation protocol is sensitive to feature \textbf{concentration} of the spurious attribute, rather than to whether the attribute has been removed from the representation. Architectural changes that concentrate rather than redistribute the attribute can produce SCR regressions even when the architecture achieves better disentanglement.

\section{Autointerp drill-down}
\label{app:autointerp}

On BatchTopK, the bottleneck reduces the mean autointerp score from .869 to .857 ($-$1.2 pp), while Matryoshka is unchanged (.865 vs.\ .865). We run a per-feature drill-down to localize this regression, averaging per-latent scores across 3 random seeds (each sampling 1000 features from the same dictionary).

Table~\ref{tab:autointerp_firerate} bins features by fire rate and compares mean autointerp score between $r{=}0$ and $r{=}24$ on BatchTopK. The regression is small and distributed across bins rather than concentrated in any single fire-rate population. The largest per-bin delta ($-$1.8 pp) occurs in the [0.01, 0.05] range. The two residual dense latents with fire rate $> 0.3$ in $r{=}24$ score .464, but represent only 2 of 2794 sampled features.

\begin{table}[h]
  \caption{Mean autointerp score by fire-rate bin on BatchTopK ($r{=}0$ vs.\ $r{=}24$, 3-seed average). The $-$1.2 pp overall regression is distributed across bins. $n$ = number of unique features sampled in that bin.}
  \label{tab:autointerp_firerate}
  \centering
  \small
  \begin{tabular}{rrrrrr}
    \toprule
    Fire-rate range & $n_{r{=}0}$ & Score$_{r{=}0}$ & $n_{r{=}24}$ & Score$_{r{=}24}$ & $\Delta$ \\
    \midrule
    $[0, 0.001)$    & 1587 & .915 & 1384 & .911 & $-$.004 \\
    $[0.001, 0.005)$ &  901 & .831 & 1070 & .828 & $-$.004 \\
    $[0.005, 0.01)$  &  193 & .738 &  253 & .737 & $-$.001 \\
    $[0.01, 0.05)$   &   89 & .733 &   83 & .715 & $-$.018 \\
    $[0.05, 0.1)$    &    6 & .583 &    1 & .786 & +.202 \\
    $[0.1, 0.3)$     &    5 & .714 &    1 & .786 & +.071 \\
    $[0.3, 1.0]$     &    0 & ---  &    2 & .464 & --- \\
    \bottomrule
  \end{tabular}
\end{table}

The score histogram reveals the mechanism. The number of features scoring in the top bin ($[0.93, 1.0]$) drops from 1428 to 1338 ($-$90), accounting for most of the mean shift. Features newly activated by the bottleneck (252 features dead in $r{=}0$ but alive in $r{=}24$) score slightly below the $r{=}24$ average (.843 vs.\ .857), contributing a small additional drag. The overall pattern is consistent with the bottleneck redistributing dictionary capacity (freeing 134 previously dead slots, shifting fire-rate distributions) in a way that marginally reduces the fraction of near-perfect explanation scores without degrading any particular feature population.

Evaluating only the top-200 features by fire rate yields .699 ($r{=}0$) vs.\ .701 ($r{=}24$), confirming that the highest-fire-rate features are not the source of regression. On Matryoshka, the top-200 evaluation drops from .731 to .705, despite the overall random-1000 score being unchanged, suggesting that Matryoshka's group structure redistributes high-fire-rate content differently.

\section{Dense latent classification details}
\label{app:dense_details}

We reimplemented five of the six automated classifiers described by \citet{Sun2025a}, including position tracking (Spearman $|\rho| > 0.4$ between decoder projection and distance to sentence/paragraph/context boundaries), PCA alignment ($|\text{cos}(\mathbf{W}_\text{dec}, \text{PC}_1)| > 0.75$), nullspace overlap ($\alpha_{10} > 0.2$), alphabet bias ($\geq$90\% shared initial letter among top-100 logit tokens), and meaningful-word part-of-speech classification (AUC $> 0.75$ on Brown Corpus noun/verb/adjective/adverb). We omit context binding, which requires LM steering.

In the BatchTopK $r{=}0$ baseline, these tests classify 7 of 25 dense latents as 2 position tracking, 1 PCA-aligned, and 4 part-of-speech. Nullspace and alphabet latents are absent, consistent with \citet{Sun2025a}'s finding that these categories emerge predominantly in late layers (L$\geq$20) rather than at L12. In BatchTopK $r{=}24$, only 4 dense latents remain, with 0 classified by the five tests. On Matryoshka $r{=}0$, group $g_0$ (512 atoms) harbors 67 of 69 total dense latents, of which 13 are classified (2~position, 2~PCA, 9~part-of-speech). Adding the bottleneck reduces $g_0$ to 41 dense latents with 0 classified.

To verify that these dense latents track the same structure the bottleneck absorbs, we examine where their decoder weights project in PC space. In the BatchTopK $r{=}0$ baseline, the 25 dense latents project predominantly onto the top-50 principal components, with a median top-PC index of 8 and 13 of 25 concentrated in the top-10 PCs. In BatchTopK $r{=}24$, only 4 dense latents remain, and 3 of these project onto PC index 0 with max cosine $> 0.70$, suggesting that these residual dense features track the very top of the variance spectrum that the rank-24 bottleneck absorbs almost but not entirely.

The connection between dense latents and the scaffold component is tight. Of the 25 dense latents in BatchTopK $r{=}0$, 25 of 25 fall in the top decile of decoder weight alignment with the scaffold component ($\alpha > 0.3$), and 19 of 25 have $\alpha > 0.7$. These features fire at rates of 10--89\%, consistent with tracking a component present. When counting features with high decoder weight alignment to the top-24 outlier-dimension component ($\alpha_\text{outlier} > 0.3$), the count drops from 787 in BatchTopK $r{=}0$ to 99 in BatchTopK $r{=}24$, indicating that the bottleneck relieves approximately 688 dictionary slots from tracking this low-rank content.

\section{Causal intervention protocol}
\label{app:causal_protocol}

All causal interventions in Section~\ref{sec:causal} use 200 sequences (1024 tokens each) drawn from \texttt{monology/pile-uncopyrighted}. For each sequence, we run a forward pass through Gemma-2-2B and intervene on the layer-12 residual stream activation before continuing to the model head. Cross-entropy is computed over all next-token predictions in each sequence and averaged across the evaluation set. The baseline (unmodified activations) achieves CE $= 1.90$.

%%%%%%%%%%%%%%%%%%%%%%%%%%%%%%%%%%%%%%%%%%%%%%%%%%%%%%%%%%%%

% \newpage
% \input{checklist.tex}

\end{document}